\documentclass[11pt]{article}

\usepackage{url} 
\usepackage[utf8]{inputenc} 
\usepackage{booktabs} 
\usepackage{graphicx}
\usepackage[figuresright]{rotating}
\usepackage[T1]{fontenc}
\usepackage{multirow}
\usepackage{tabularx}
\usepackage{makecell}
\usepackage{adjustbox}
\usepackage{siunitx}
\usepackage{geometry}

\usepackage{authblk}
\usepackage{lmodern} 
\usepackage[T1]{fontenc} 
\usepackage{bm} 
\usepackage{helvet} 
\usepackage{amsmath}
\usepackage{hyperref}

\usepackage{multirow}
\usepackage{wrapfig}
\usepackage{xspace}
\usepackage[capitalise,noabbrev,nameinlink]{cleveref}
\crefformat{equation}{#2Equation~#1#3}
\usepackage{pgfplots,pgfplotstable} 
\usepackage{subcaption}
\usepackage{color}

\usepgfplotslibrary{colorbrewer,groupplots}
\usetikzlibrary{positioning}
\pgfplotsset{compat=1.18}
\definecolor{visible-blue}{rgb}{0.286, 0.525, 0.910}
\definecolor{tabfirst}{rgb}{1, 0.7, 0.7} 
\definecolor{tabsecond}{rgb}{1, 0.85, 0.7} 
\definecolor{tabthird}{rgb}{1, 1, 0.7} 

\newcommand{\OOM}[1]{-}
\usepackage{tocloft}

\begin{document}
\thispagestyle{empty}

\title{\Large \bf LandMarkSystem Technical Report}

\author[1,2]{Zhenxiang Ma}
\author[1]{Zhenyu Yang}
\author[1]{Miao Tao}
\author[1]{Yuanzhen Zhou}
\author[1]{Zeyu He}
\author[1]{Yuchang Zhang}
\author[1]{Rong Fu}
\author[1]{Hengjie Li\thanks{Corresponding author (lihengjie@pjlab.org.cn).}}

\affil[1]{Shanghai Artificial Intelligence Laboratory, Shanghai, China}
\affil[2]{Shanghai Jiao Tong University, Shanghai, China}
\date{}

\maketitle

\begin{abstract}
3D reconstruction is vital for applications in autonomous driving, virtual reality, augmented reality, and the metaverse. Recent advancements such as Neural Radiance Fields (NeRF) and 3D Gaussian Splatting (3DGS) have transformed the field, yet traditional deep learning frameworks struggle to meet the increasing demands for scene quality and scale. This paper introduces LandMarkSystem, a novel computing framework designed to enhance multi-scale scene reconstruction and rendering. By leveraging a componentized model adaptation layer, LandMarkSystem supports various NeRF and 3DGS structures while optimizing computational efficiency through distributed parallel computing and model parameter offloading. Our system addresses the limitations of existing frameworks, providing dedicated operators for complex 3D sparse computations, thus facilitating efficient training and rapid inference over extensive scenes. Key contributions include a modular architecture, a dynamic loading strategy for limited resources, and proven capabilities across multiple representative algorithms. This comprehensive solution aims to advance the efficiency and effectiveness of 3D reconstruction tasks.To facilitate further research and collaboration, the source code and documentation for the LandMarkSystem project are publicly available in an open-source repository, accessing the repository at: \hyperlink{LandMark}{https://github.com/InternLandMark/LandMarkSystem}.

\end{abstract}
\begin{figure}[!h]
\centering
\includegraphics[width=0.8\columnwidth]{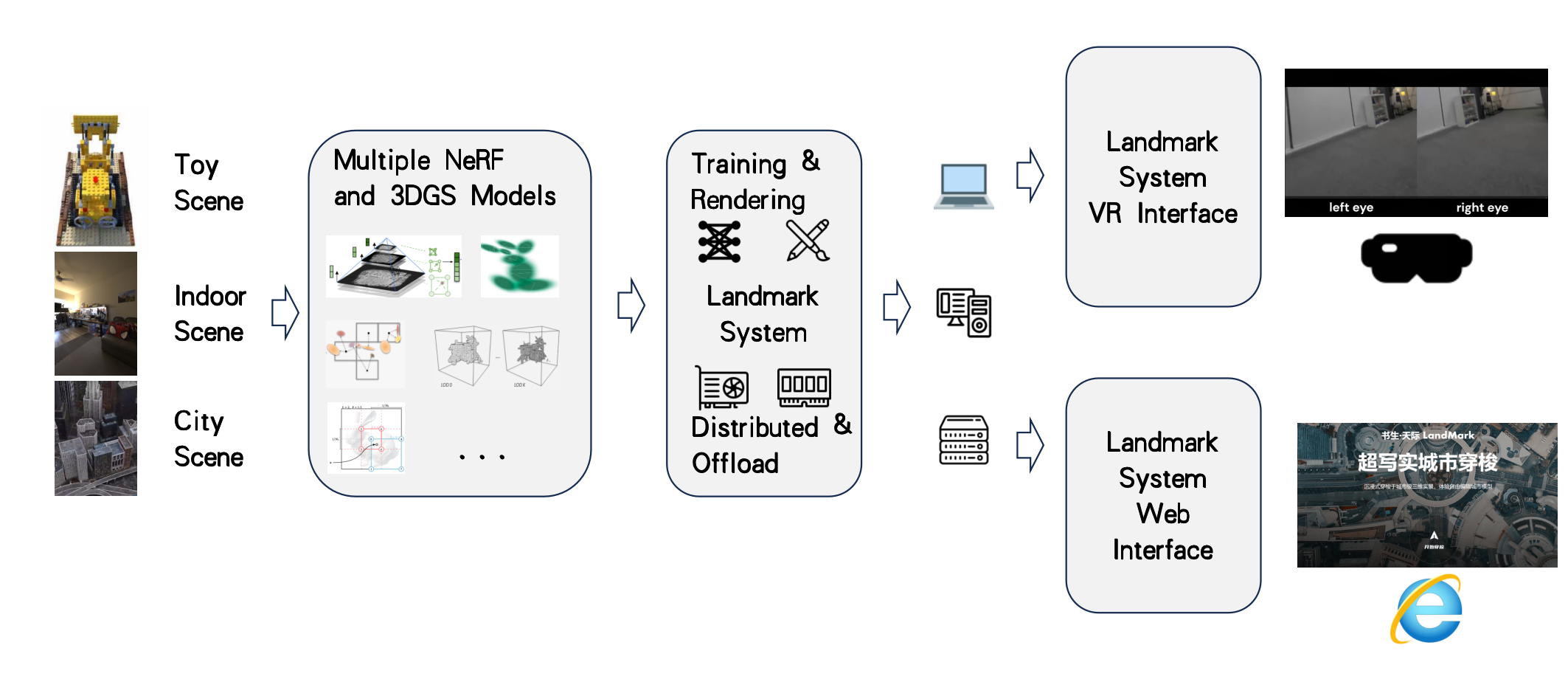} 
\caption[Overview]{LandMarkSystem supports various mainstream NeRF and 3DGS reconstruction algorithms. Through meticulous system optimization, it seamlessly adapts to scenes ranging from individual objects to entire cities and scales from single consumer-grade GPU to distributed clusters, delivering a smooth interactive experience on Web and VR platforms.}
\label{fig_model_components}
\end{figure}

\clearpage
\tableofcontents
\clearpage
\section{Introduction}
3D reconstruction plays a crucial role in numerous fields, including autonomous driving, virtual reality (VR), augmented reality (AR), the metaverse, etc. Neural Radiance Fields (NeRF)~\cite{Mildenhall2020NeRF} have introduced a novel approach for 3D reconstruction. The following 3D Gaussian Splatting (3DGS) ~\cite{Kerbl20233DGS} replaces the implicit representation of NeRF, achieving increased speed in the reconstruction process. As the demand for finer scene quality and larger scene scale increases, new algorithm structures are emerging continuously. As a result, the conventional deep learning computing framework is no longer sufficient to support rational speed in the reconstruction and inference of different NeRF and 3DGS.

The enhancement of quality and the expansion of scale correspond to a surge in the complexity of scene information, thereby presenting a formidable challenge to the accommodation and reproductive capacities of the 3D reconstruction model. Scaling the model parameters is an intuitive way to increase the capacity of the model, but it is unable to solve the fundamental problem. For NeRF, more parameters will not necessarily improve the capacity of the model, because the features that can effectively represent the scene in a continuously distributed neural field always tend to be discretely distributed. For 3DGS, more parameters can increase the capacity of the model, but the discrete distribution of the Gaussians significantly increases computational and storage overhead. In this paper, we introduce LandMarkSystem, a computing framework for multiple-scales scene reconstruction and rendering. It supports a variety of different NeRF and 3DGS algorithm structures through a componentized model adaption layer, and expands the capacity and adaptability of the model based on the scene multi-block/branch patition strategy. It natively supports distributed parallel computing and model parameter offload optimization in distributed adaption layer, provides a highly optimized operator library for bottlenecks in the computation stage, and can achieve efficient training and rapid inference of scene geometries up to hundreds of square kilometers.

The essence of 3D reconstruction is to learn 3D spatial information, which makes classic computing frameworks which focus on one-dimensional or two-dimensional information such as text and images weak on related tasks, such as Pytorch\cite{10.5555/3454287.3455008}, TensorFlow\cite{10.5555/3026877.3026899}, etc. These deep learning frameworks usually have very strong versatility and scalability, which are sufficient to implement the basic computing process of algorithms such as NeRF and 3DGS, but they are not able to meet the needs of development in terms of ease of use, and lack dedicated operators to support complex sparse computation in 3D space, resulting in low training and reasoning efficiency. In order to improve ease of use in a targeted manner and facilitate experimental research on algorithm structures, a series of computing frameworks for 3D reconstruction and rendering, such as NeRFStudio\cite{nerfstudio}, Kaolin-wisp\cite{KaolinWispLibrary}, etc., and dedicated operator libraries for sparse calculations in 3D space, such as Nerfacc\cite{10376856}, have been proposed. These works have built a strong community influence and quickly promoted the related work of NeRF and 3DGS, ut the overall training and reasoning efficiency is still limited by image resolution, scene scale, model parameter quantity, etc. LandMarkSystem provides a novel solution from the perspective of computing systems. It is compatible with a variety of algorithm structures through modular model components and data tools, facilitating the development work of researchers. Based on the scene block strategy, LandMarkSystem supports efficient training and real-time rendering (>30FPS) of large-scale data sets and large-scale scenes through a native, distributed parallel engine with integrated parameter offload and dedicated sparse computation operators, providing an end-to-end complete pipeline.

Our main contributions include:
\begin{itemize}
\item Modular model adaptation layer, including algorithm components and management components, which is capable to implement various NeRF and 3DGS model structures
\item Distributed adaptation layer, including distributed parallel computing framework APIs, and converters that automatically convert the aforementioned algorithm components into parallel forms
\item Dynamic loading strategy based on model parameter offload, achieving training and rendering of large-scale scenes on limited computing resources
\item Support for multiple application scenarios, thanks to the systematic architecture, realizing end-to-end support from reconstruction to real-time rendering
\item The system's training and rendering capabilities have been tested on a variety of representative NeRF and 3DGS algorithms
\end{itemize}



\section{Preliminaries}
The LandMarkSystem revolves various algorithms based on NeRF and 3DGS. These algorithms form the core for implementing training and rendering acceleration strategies within the system. Here, we provide a brief overview of the two main algorithm types mentioned above:

\subsection{Neural Radiance Fields}
NeRF uses weights learned through a multi-layer perceptron (MLP) to model a scene and represent it as a continuous volume distribution with color. The distributions represent the probability of light being blocked or transmitted in volumetric rendering \cite{468400}. The MLP network takes three-dimensional position \textbf{x} = ($x$, $y$, $z$) and unit-normalized direction \textbf{d} = ($d_x$, $d_y$, $d_z$) as input, then outputs the corresponding density \textbf{$\sigma$} and color \textbf{c} = ($r$, $g$, $b$). When outputting a 2D image for a given position and direction, NeRF assumes a camera ray r($t$) = \textbf{o} + $t$\textbf{d} is emitted from the camera's projection center \textbf{o}, passes through the given pixel on the image plane along direction \textbf{d}, and approximates the pixel's rendered color through numerical integration\cite{7780814}:

\begin{equation}
\hat{C}(\textbf{r}) = R(\textbf{r}, \textbf{c}, \textbf{$\sigma$}) = \sum_k T_k (1 - \exp(- \textbf{$\sigma$}(t_{k+1} - t_k))) \textbf{c}(t_k),
\end{equation}
\begin{equation}
T_k = \exp(-\sum_{k'<k} \textbf{$\sigma$} (t_{k'+1} - t_{k'}))
\end{equation}

Here, $R$ represents the volume rendering result of the density \textbf{$\sigma$} and color  \textbf{$\sigma$}  obtained along  \textbf{$\sigma$}, where  \textbf{$\sigma$}($t$) and  \textbf{c}($t$) are the corresponding values at a specific point r($t$). The interval between the pre-defined near plane $t_n$ and far plane $t_f$ of the camera is sampled to obtain the set \textbf{t} based on the sampling strategy. NeRF obtains the values of \textbf{$\sigma$}($t$) and \textbf{c}($t$) as follows. For $\forall t_k \in t$ , obtain its position coordinates \textbf{x} = r($t_k$) = \textbf{o}+$t_k$ \textbf{d}  on the ray, and the corresponding direction \textbf{d}  can also be obtained. NeRF does not directly use \textbf{x}  and \textbf{d} as MLP inputs, but the encoded results $\gamma _\textbf{x}(\textbf{x})$ and $\gamma_\textbf{d} (\textbf{d})$ instead. If the length of the encoded sequence is $S$, the positional encoding function form is:

\begin{equation}
\begin{split}
\gamma(x)= (sin(2^0x),sin(2^1x),sin(2^2x),...,sin(2^{S-1}x),
\\ cos(2^0x),cos(2^1x),cos(2^2x),...,cos(2^{S-1}x))
\end{split}
\end{equation}

Next, after passing through the MLP with parameters \textbf{$\theta$} = ($\theta_1$,$\theta_2$), the corresponding density \textbf{$\sigma$}($t$) and color \textbf{c}($t$) are obtained. Due to the strategy of using staged input and output, the calculation of the density \textbf{$\sigma$}($t$) is independent of the given direction, while the calculation of the color \textbf{c}($t$) is related to both the given position and direction:

\begin{equation}
[ \textbf{$\sigma$}(t),\textbf{u}(t)]= MLP_{\theta_1} (\gamma_\textbf{x}(\textbf{x})),
\end{equation}
\begin{equation}
\textbf{c} (t)= MLP_{\theta_2} (\gamma_\textbf{d} (\textbf{d} ),\textbf{u}(t))
\end{equation}

Performing global direct sampling in space would take a lot of time to achieve convergence. Therefore, NeRF optimizes two MLP networks simultaneously and names them the coarse model and fine model according to different sampling strategies. The coarse model performs uniform sampling and density prediction,
determines the sampling points for the fine model by PDF sampling, thus quickly excluding the space in the scene that does not contribute to rendering. The two models are optimized by minimizing the following loss:

\begin{equation}
\sum_\textbf{r}((\vert\vert C^*(r)-\hat{C_c}\vert\vert_2^2)+(\vert\vert C^*(r)-\hat{C_f}\vert\vert_2^2))
\end{equation}

Where  $C ^* (r)$ is the pixel color corresponding to the input image and ray r , $\hat{C_c}$ and $\hat{C_f}$ are the prediction of pixel colors output by the coarse model and fine model, respectively.

\subsection{3D Gaussian Splatting}
3DGS represents the structure and color of a scene using a series of anisotropic 3D Gaussians. Unlike the traditional volumetric rendering approach utilized by NeRF, 3DGS employs differentiable rasterization with opacity blending ($\alpha$-blending). The reconstruction process in 3DGS typically starts from sparse point clouds, which can be randomly generated or obtained through Structure-from-Motion (SfM)\cite{964490} applied to the scene images. In a sparse point cloud, the position coordinates of each element serve as the mean $\mu$, generating the corresponding 3D Gaussian. 

\begin{equation}
G(x) = \exp(-(x-\mu)^T\Sigma^{-1}(x-\mu)/2)
\end{equation}

Where $x$ represents any position in the scene space, and $\Sigma$ denotes the covariance matrix of the 3D Gaussian. $\Sigma$ can be decomposed into a rotation matrix $R$  and a scaling matrix $S$ to maintain its positive definiteness.

\begin{equation}
\Sigma = RSS^TR^T
\end{equation}

In addition to the mentioned attributes, each 3D Gaussian also includes a set of color coefficients encoded using spherical harmonics and an opacity value $\alpha$. These are used for subsequent opacity blending operations during rasterization process. While rendering, the 3D Gaussians are first projected onto screen space using the EWA algorithm\cite{964490} and transformed into 2D Gaussians—a process commonly referred to as $Splatting$. Subsequently, the rasterizer performs depth sorting and opacity blending based on the 2D Gaussians. This rasterization process operates at the granularity of tiles, where each tile corresponds to a partition of pixels in screen space, enhancing computational parallelism and efficiency. Finally, the color of a single pixel is computed as follows:

\begin{equation}
\hat{C}(x')=\sum_{i\in N}c_i\sigma_i\prod_{j<i}(1-\sigma_j),
\end{equation}
\begin{equation}
\sigma_i = \alpha_iG'_i(x')
\end{equation}

Where $x'$ represents the position of a pixel in screen space, and $N$ denotes the number of 2D Gaussians involved in rendering the corresponding pixel. Leveraging the differentiate property of the rasterizer, all parameters of the 3D Gaussians can be optimized through an end-to-end rendering process integrated with training, allowing for scene reconstruction under supervised learning

\section{System Design}
\begin{figure}[!h]
\centering
\includegraphics[width=0.74\columnwidth]{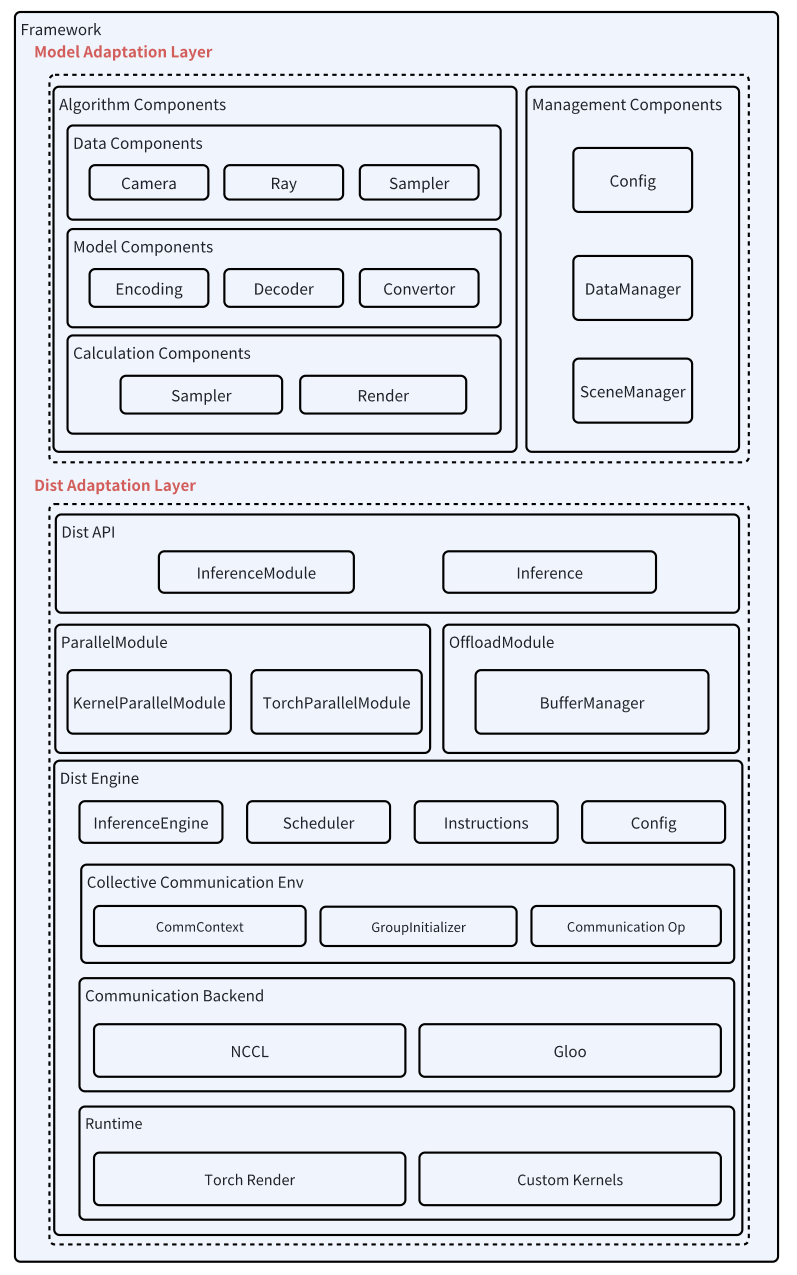}
\caption[Framework]{LandMarkSystem Framework: The design of the LandMarkSystem is bifurcated into the Model Adaptation Layer and the Distributed Adaptation Layer. The Model Adaptation Layer equips the system with the capability to accommodate a variety of NeRF and 3DGS algorithms, ensuring compatibility and versatility. The Distributed Adaptation Layer, on the other hand, empowers the models with parallel processing capabilities. This tiered architecture not only facilitates the training and rendering of conventional small-scale object-level scenes but also seamlessly scales to address ultra-large-scale urban scenarios.}
\label{fig_sys_design}
\end{figure}

In the context of commonly employed NeRF and 3DGS algorithms, there exists a degree of structural similarity or generality, such as input module for different datasets, rasterizer for 3DGS based methods, etc. Reusing these common elements can enhance the design efficiency for algorithm developers. Moreover, distributed model parallelism can be achieved through a component-based architecture. In our design, the LandMarkSystem establishes a modular framework that facilitates a unified workflow supporting various algorithms. With a componentized structure and a hierarchical system design, algorithmic models can be readily scaled to multi-GPU parallelism. Furthermore, through dynamic loading, the system enables the rendering of an infinite area on a single GPU. Sec.~\ref{System} outlines the system with the Model Adaptation Layer and the Distributed Adaptation Layer. Sec.~\ref{Model Adaptation} details the structure of the Model Adaptation Layer. Sec.\ref{Distributed Layer} discusses the interface api and different modules with the hidden Distributed Adaptation Layer. Finally, Sec.\ref{system_application} summarizes the system capabilities and application scenarios.

\subsection{System Design}
\label{System}
As illustrated in Fig.~\ref{fig_sys_design}, the overall architecture of the LandMarkSystem is structured around the Model Adaptation layer and the Dist Adaptation layer. The Model Adaptation layer provides a modular structure for managing model, algorithm components and management components. Within this layer, the algorithm component comprises sub-components such as data components, model components, and calculation components. Management components handle data management, scene setup, and global configurations, respectively. The model component offers a framework for both sequential and parallel algorithm structures. Through the components conversion interface, it automatically translates the constructed sequential algorithms into parallel ones. Besides, it could also apply kernel optimization into the components.

The Distributed Adaptation Layer operates seamlessly in the background from the user's perspective. It requires only the specification of a parallel environment in the configuration file. This triggers the InferenceModule, which selects between ParallelModule or OffloadModule to adapt to large-scale scenes. The Distributed Engine is responsible for distributed rendering, encompassing processes related to communication and the registration of specialized operators.

\subsection{Model Adaptation Layer}
\label{Model Adaptation}
\begin{figure}[!h]
\centering
\includegraphics[width=0.9\columnwidth]{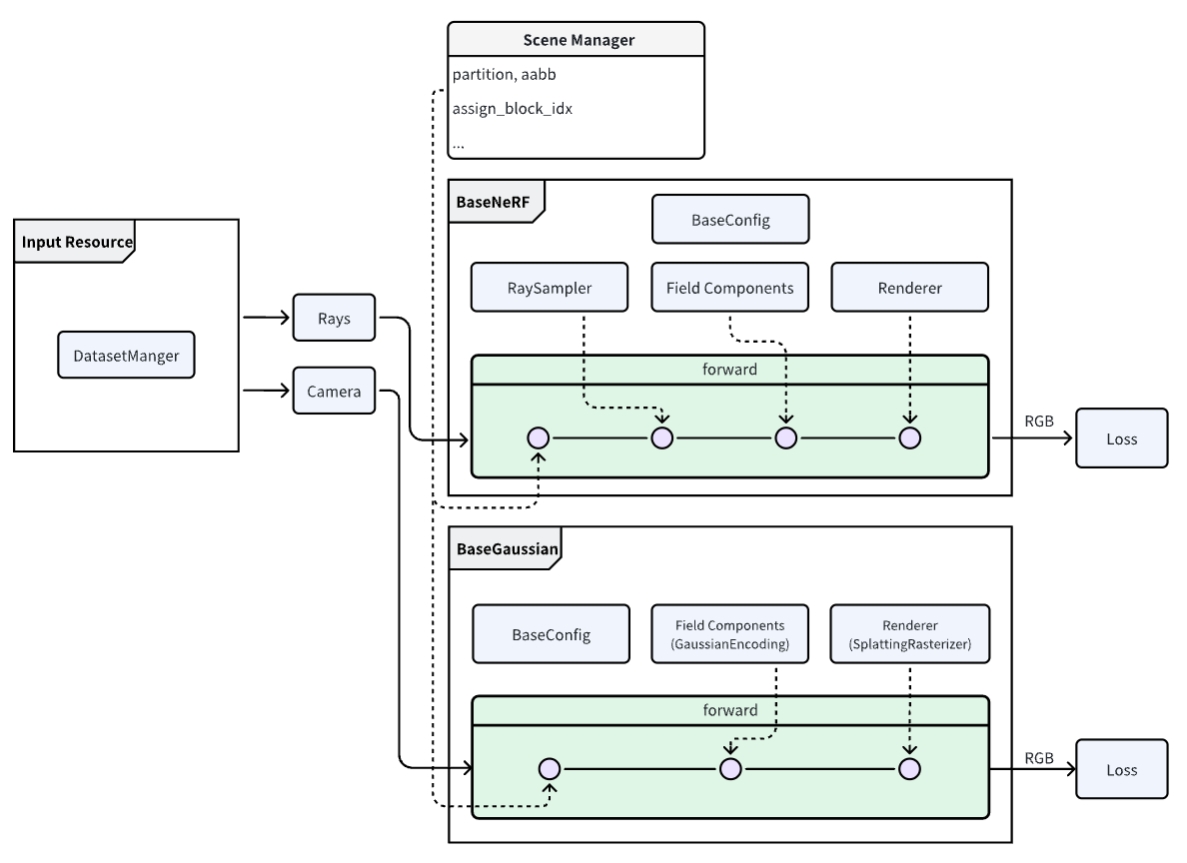} 
\caption[Pipeline]{LandMarkSystem Pipeline: The operational flow of the algorithm involves selecting between Rays or Camera inputs during the input stage, depending on the model. These inputs then enter the pipelines of the NeRF class and the 3DGaussian class, respectively, for training or rendering.}
\label{fig_components_interact}
\end{figure}

The Model Adaptation Layer of the LandMarkSystem aims to achieve two primary goals: efficient modularization, which provides a unified operational workflow for multiple algorithms, and high encapsulation, which hides the system’s complex operations and interfaces, exposing only the functionalities required by users to facilitate ease of use and secondary development. Drawing from the characteristics of existing NeRF and 3DGS algorithms, we categorize the algorithm components into three types: data components, model components, and calculation components. In addition to the algorithm components, we also have a management component responsible for managing scenes and datasets. These components interact to construct a complete and unified process, as shown in Fig.~\ref{fig_components_interact}.

According to the different rendering computation principles, the models are divided into two types: BaseNeRF and BaseGaussian. The input of the entire pipeline is generated by the DatasetManager according to different types of models. Rays are used for BaseNeRF and Camera for BaseGaussian. The SceneManager is responsible for managing the entire scene and its block structure. Subsequently, different algorithms will calculate RGB and loss based on their algorithm components. For BaseNeRF, the workflow mainly consists of three parts: the data component (Rays and Samples), the model component (FieldComponents), and the calculation component (Sampler and Renderer). For Gaussian models, since they do not need to sample rays, they only contain the model component (FieldComponents) and the calculation component (Renderer). We will introduce these components in the following subsections.

\subsubsection{Algorithm Components} The algorithm components consists of three type components: data components, model components and calculation components. This section will further introduce them in detail.
\paragraph{\textbf{Data Components}}
Data components aggregate similar data and functions, encapsulating underlying communication operations. They are categorized into three types:
$$\textbf{Cameras} \rightarrow pose2ray \rightarrow \textbf{Rays} \rightarrow Sampler \rightarrow \textbf{Samples}$$
The Cameras component manages camera data, storing parameters and providing the pose2ray interface for subsequent rendering pipeline input. The Rays component handles ray information, which is passed to the Sampler to generate Samples. The Samples component offers the ability to access and communicate data for a batch of sampling points across different scene blocks. It not only manages the attributes of Samples but also encapsulates computations directly related to the Samples class and manages rendering results such as $RGB$ and $\sigma$. For 3DGS based on rasterization rendering, it can skip the Rays and Samples generation and directly use Camera as input.

\paragraph{\textbf{Model Components}}

Model components constitute the core of the model. The current system supports two types of models: NeRF and 3DGS. For NeRF, the input is the Samples and the output is the $RGB\sigma$ of the samples. For 3DGS, the input is the Camera and the output is the image. Model components, including encoding and decoding, encapsulate the specifics of the algorithms, considering classic computational characteristics. The typical pipeline involves extracting features using a specific encoder from input data, then decoding the features and rendering them into images.

Encoding is the scene representation module. Large scenes require encoding representation with a large number of parameters. Large model parallelism is primarily achieved through the Encoding components, reducing the parameter load on a single GPU. 
The Decoder is the decoding part of the model, generally with fewer parameters and not requiring model parallelism.

\begin{figure}[!h]
\centering
\includegraphics[width=0.9\columnwidth]{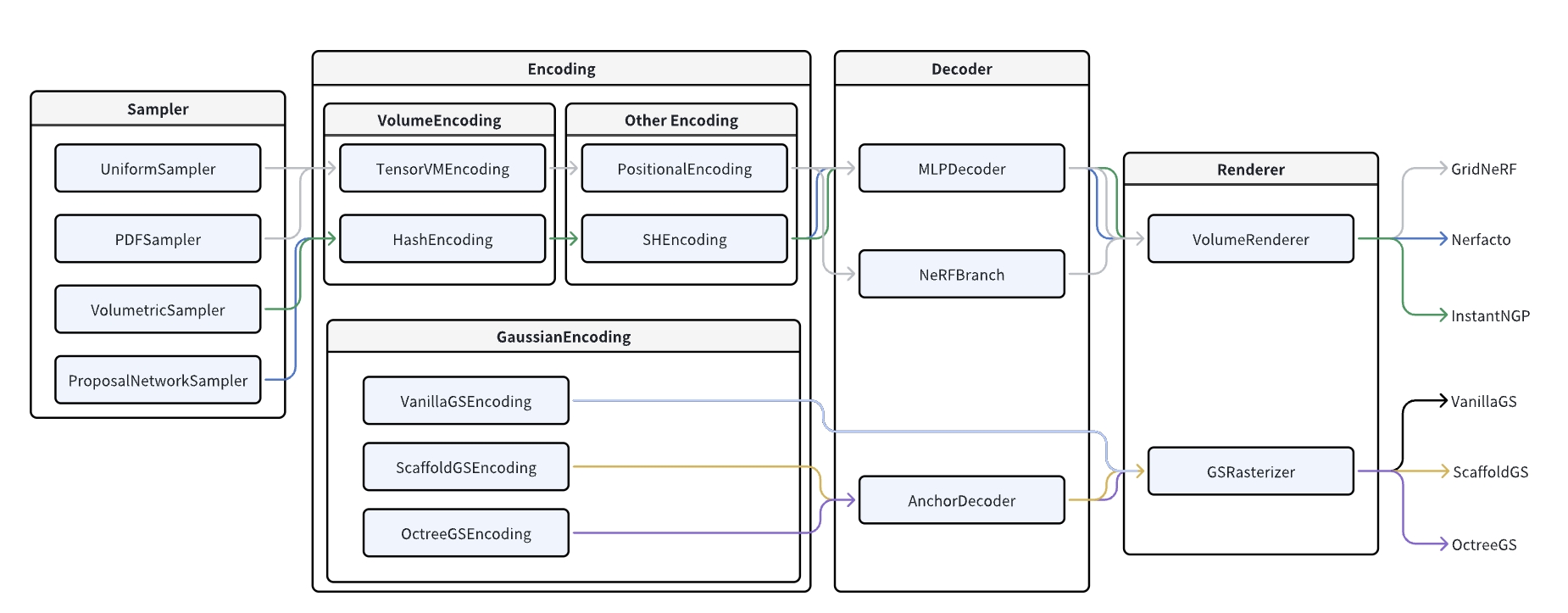} 
\caption[Algorithm]{LandMarkSystem Algorithm: Examples of modularizing algorithm components to build different algorithms}
\label{fig_model_components}
\end{figure}

Fig~\ref{fig_model_components} illustrates the construction of algorithms through components in LandMarkSystem. Currently, the system has achieved complete implementations of InstantNGP, nerfacto, GridNeRF, 3DGS, scaffoldGS, and OctreeGS. VolumeEncoding and GaussianEncoding are different implementations of Encoding components. MLPDecoder, NeRFBranch and AnchorDecoder are all implementations of Decoder components. They all belong to the definition of Model Components. Sampler and Renderer are both belong to the calculation components, which we will introduce in the following paragraph.

\paragraph{\textbf{Calculation Components}}
Calculation components can be regarded as converters between different data components. For NeRF, the process involves several stages:
$$ \text{Rays} \rightarrow \textbf{Sampler} \rightarrow \text{Samples} \rightarrow \text{NeRF} \rightarrow \text{Samples' RGB}\sigma \rightarrow \textbf{Renderer} \rightarrow \text{Rays' RGB}\sigma $$
The \textbf{Sampler} component converts Rays into Samples, and the \textbf{Renderer} converts $Samples' RGB\sigma$ into $Rays' RGB\sigma$. Both are crucial calculation components. For 3DGS, the calculation components are defined differently:
$$ \text{Cameras} \rightarrow \text{Gaussians Primitives} \rightarrow \textbf{Renderer (GSRasterizer)} \rightarrow \text{Image} $$
Here, \textbf{GSRasterizer} serves as the implementation of the 3DGS calculation component \textbf{Renderer}, which can directly generate images from the input Gaussian Primitives.

\subsubsection{Management Components} 
The management components consist of SceneManager and DatasetManager.
The SceneManager component manages scene information. Different from object-level reconstruction, large-scale scenes have a vast number of parameters. To reconstruct such scenes, the system relies on local processing, necessitating the ability to divide and manage small regions for parallel processing or dynamic loading. SceneManager maintains global scene information, provides location indices for Sample points and sub-models within the scene, and maintains a mapping table between blocks and sub-models to facilitate spatial region and local sub-model queries. It also offers a filter for Samples, selecting valid sample points for sub-models.

The DatasetManager component provides a unified source of input data for the model, including various sources such as training/test datasets (Dataset), real-time rendering terminal poses (Pose), and fixed trajectories (trajectory files). The dataloader enables reading from different formats including Colmap dataset and Blender format dataset.

\subsection{Distributed Adaptation Layer}
\label{Distributed Layer}
The design of the distributed layer is mainly to hide the internal implementation of the system from users, so that users can directly apply the various optimization methods provided by the system through simple API calls. The distributed layer is mainly composed of the following parts: \textbf{Distributed API} used to build a distributed inference engine.
\textbf{ParallelModule} is used to parallelize the defined sequential algorithm model. Through \textbf{OffloadModule}, the algorithm model is split and stored in GPU and CPU memory and the rendering parameters are dynamically loaded and offload between them.
\textbf{Distributed Engine} responsible for building and managing the distributed environment including NCCL and Gloo, and providing two runtime environments Torch runtime and Custom Kernel Runtime.

\subsubsection{Distributed API} 
\begin{figure}[h]
    \centering
    \begin{subfigure}{0.48\textwidth}
        \centering
        \includegraphics[width=\linewidth]{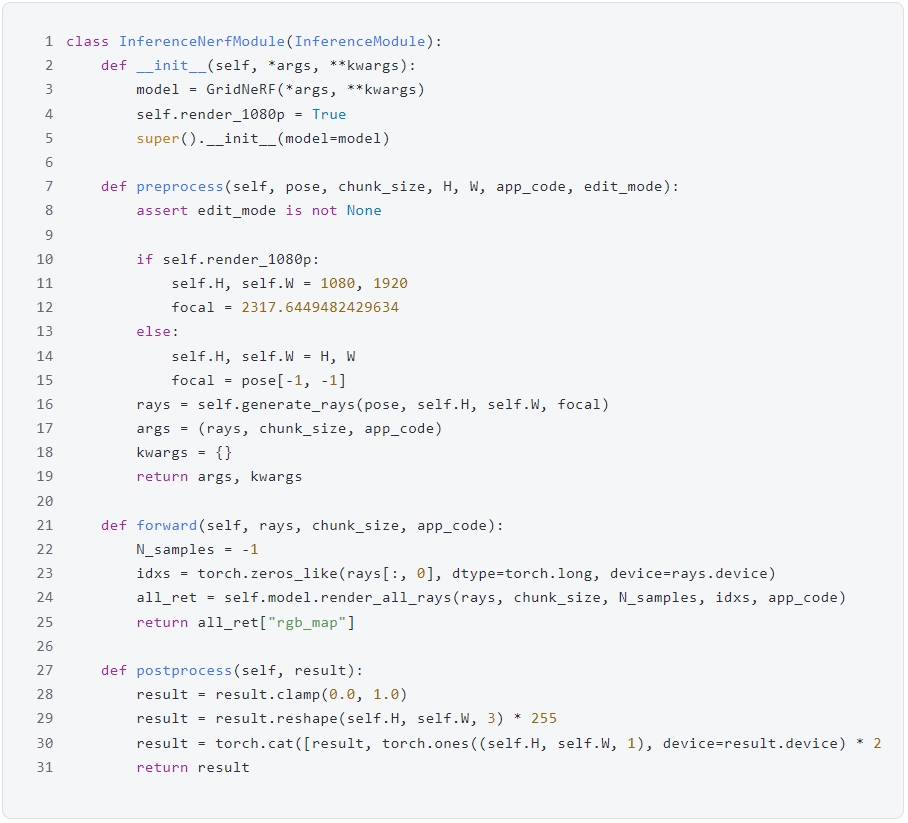}
        \caption{InferenceModule Demo}
        \label{fig:inference_module}
    \end{subfigure}
    \hfill
    \begin{subfigure}{0.45\textwidth}
        \centering
        \includegraphics[width=\linewidth]{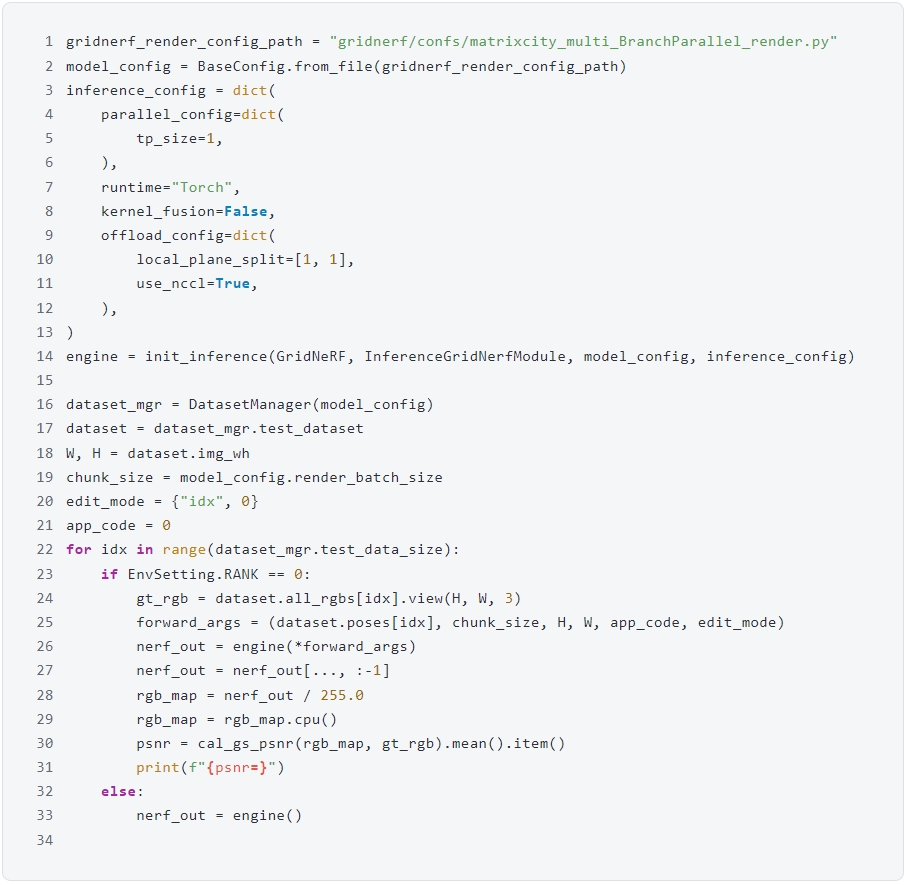}
        \caption{InitInference Demo}
        \label{fig:init_inference}
    \end{subfigure}
    \caption{Distributed API}
    \label{fig:combined}
\end{figure}

The features of the entire distributed layer are provided to users through the distributed API. There are two main interface API functions: InferenceModule and InitInference. We divide the process of rendering an image into three stages: \textbf{preprocess}, \textbf{forward} and \textbf{postprocess}. Data preparation and transformation mainly in the preprocess stage. For instance, converting pose to ray could be happened in this stage for NeRF. The forward stage is mainly the encapsulation of the forward function in the algorithm model. Postprocess is responsible for converting the results of the forward model into the output format we expect. The specific implementation logic of the above three stages and algorithm models are encapsulated and defined in the InferenceModule, and then handed over to the InitInference for initialization. A typical implementation of InferenceModule for NeRF could be seen in Fig~\ref{fig:inference_module}.

Fig~\ref{fig:init_inference} shows the usage of InitInference API. The input of the Inference API is the definition class of the algorithm model, the definition class of the InferenceModule and the configuration of the algorithm model and inference engine. The return of the API is an initialized engine which could be used by users directly. 
The input of initialized engine is the preprocess input of InferenceModule, and the output is the postprocess output. When there is no need to do any processing on the input and output of the algorithm model, user can skip defining the InferenceModule and only give the class of algorithm model to InitInference. At this time, the input and output of the initialized engine are both the input and output of the algorithm model forward.

The configurations of the algorithm model and the inference engine are parameters that must be defined to initialize the engine. The algorithm model configuration is used to help initialize the algorithm model. It defines the structure of the algorithm model and the calculation method. The inference engine configuration is mainly used to determine the engine feature to be used. For example, if user needs to enable parallel rendering, the $tp\_size$ parameter of $parallel\_config$ need to be set. The formula relationship between data parallelism and model parallelism is as follows: 
$$world\_size=dp\_size*tp\_size$$
If user needs to enable the offload, $offload\_config$ should be specified and $local\_plane\_split$ is used to control the area loaded into the GPU memory.

\subsubsection{Components Convertor}
The conversion of algorithm models into parallel models and switching of runtime within the engine are both achieved through Components Convertor. Parallel conversion and runtime conversion are triggered by $parallel\_config$ and $runtime$ in the inference configuration, respectively. 

The model parallel partitions the encoding parameters to distributed devices. Based on different partition dimensions or rules, various partitions could be applied. LandMarkSystem provide mainly two ways for model parallel. The first scheme is \textit{Branch Parallel}, which partition the model parameters focus on the geometry positions, like (x,y) coordinate. Using the locality of viewing the 3D reconstruction scene, it is natural and similar to traditional computer graphics. Starting with a complete scene, we first partition the model into several blocks in grid. Corresponding images and sample points are assigned to blocks by the coordinate. The other options for parallel is \textit{Channel Parallel}, which divides the features of encoding part following channel dimension to different devices. This parallel method is more likely to intra-layer model parallel, which parallel the DNN models in layer's inner structure. It is worth noting that this method relies on the decomposition of encoding parameters and some encoders are probably not suitable for channel parallel.

The distributed adaptation layer offers the capability of parallel computing. Based on the parallel configuration settings, it transforms sequential models into parallel ones using a wrapping method. The system defines a component mapping table: 
$$C_{\{seq\}i} \rightarrow C_{\{parl\}i}$$ 
Where $C_{\{seq\}i}$ donates to a specific component and $C_{\{parl\}i}$ is the corresponding components in parallel mode. During the conversion, it traverses the modules in the sequential model, converts their types to parallel module according to the component mapping table and initializes them. In practical usage, users only need to configure the distributed environment settings in the Config file to automatically enable parallel computing, without the need for additional programming or development efforts.

In addition to parallel conversion, we also provide kernel conversion, which can convert our algorithm components into kernel fusion implementations. The main idea is similar to the logic of parallel component conversion, and kernel conversion is orthogonal to parallel conversion.

\subsection{Conclusion and Application}
\label{system_application}

\begin{figure}[!h]
\centering
\includegraphics[width=1.0\columnwidth]{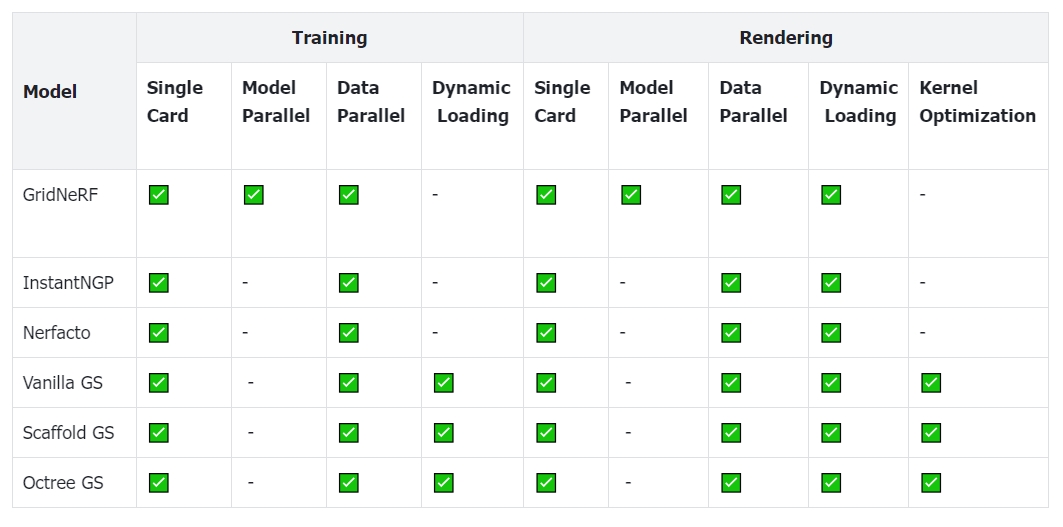} 
\caption{Feature supports in LandmarkSystem.}
\label{fig_model_components}
\end{figure}

LandMarkSystem provides different strategies for training and rendering optimization solutions for different algorithms. Model parallelism can solve the problem of insufficient video memory for large scenes. Data parallelism achieves system scalability and can provide a linear acceleration ratio between performance and the number of cards. Dynamic loading provides the ability to train and render unlimited areas on a single card. Based on LandMarkSystem, users can choose the most efficient algorithm according to different scenarios and the most appropriate optimization method according to hardware resources.

Through the componentized design of the algorithm adaptation layer and the general abstraction of the distributed adaptation layer, LandMarkSystem can efficiently support different NeRF-based and GS-based algorithms, which also enables us to support multi-scale scenes, as shown in Figure 1. Single objects, indoor scenes, urban street scenes, and even urban scenes of more than 100 square kilometers can all be trained and rendered in real time using the LandMarkSystem. In addition, we also provide a general inference engine that can be used as a backend engine for Web and VR frontend. We have thoroughly optimized the system for immersive 3DGS VR rendering and discussed it in detail in \cite{tao2025gscachegscacheinferenceframework}

\section{Optimization Methods}
\subsection{Techniques for Optimizing Training}
\subsubsection{Scaling Training Computations to City-Scale}
Training large models typically utilize  hybrid parallel strategies for scaling up, like tensor parallel, pipeline parallel, etc. For large scale 3D reconstruction training tasks, we identify the special data requirements, model architectures and computation patterns, designing multiple novel schemes towards city scale with great efficiency. While  3DGS based methods typically share similar computation patterns, the optimization techniques for 3DGS is in general. Since NeRF methods are more different from each, we choose the method\cite{Xu2023GridguidedNR} with the best performance in large scale evaluations and apply extra optimizations for it.

\paragraph{Channel Parallel}
GridNeRF uses Grid-Based scene representation, and its encoding is expressed more compactly by decomposing the feature network. For specific derivation, refer to \cite{Xu2023GridguidedNR} . In GridNeRF, there are:
\[
\mathcal{G}_{\sigma}^{n} = \bigoplus \left[ \left( \mathbf{v}_{\sigma,r}^{z} \circ \mathbf{M}_{\sigma,r}^{xy} \right) \right]_{R_{\sigma}}, \quad
\mathcal{G}_{c}^{n} = \bigoplus \left[ \left( \mathbf{v}_{c,r}^{z} \circ \mathbf{M}_{c,r}^{xy} \right) \right]_{R_{c}}
\]
To simplify the expression, we call the feature matrix M "Plane" and the feature vector v "Line". We split the plane and line along the feature dimension respectively, and evenly divide the parameters of the model encoding part to different devices, which can effectively alleviate the memory pressure of a single GPU. During forward calculation, each device uses its plane and line with some feature dimensions to calculate local features, and then performs all-gather aggregation to obtain complete features. We call this parallel training method Channel Parallel, which has good load balancing characteristics in both memory and computing. The maximum parallelism of Channel Parallel depends on the feature length of the model plane and line, which is very suitable for small-scale scenes with limited area and the number of model parameters exceeds the usage scenario of a single GPU device memory.
\begin{figure}
    \centering
    \includegraphics[width=0.5\linewidth]{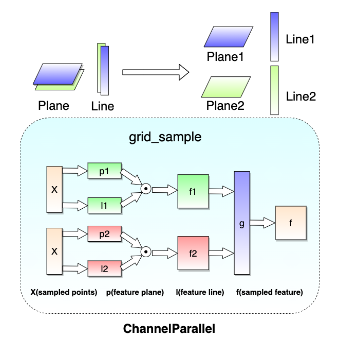}
    \caption[Channel Parallel]{Channel Parallel}
    \label{fig:enter-label}
\end{figure}
\paragraph{Branch Parallel}
We further proposed the Branch Parallel training method to meet the modeling needs of more than hundreds of square kilometers, which can theoretically achieve unlimited scale-up of modeling area and model size. As shown in the figure below, for the Encoding representation of a single super-large scene, we first divide the scene into grid-like areas on the x-y plane. For each grid area, we use an independent Encoding to represent the scene. The Encoding of each sub-grid constitutes a branch of the complete scene model. All sub-branches and the shared MLP part constitute a complete multi-branch large model. This parallel method of using an independent and complete Encoding branch for each grid area can ensure that the modeling quality is still high in each grid area while the modeling size is arbitrarily scaled up.
\begin{figure}
    \centering
    \includegraphics[width=0.5\linewidth]{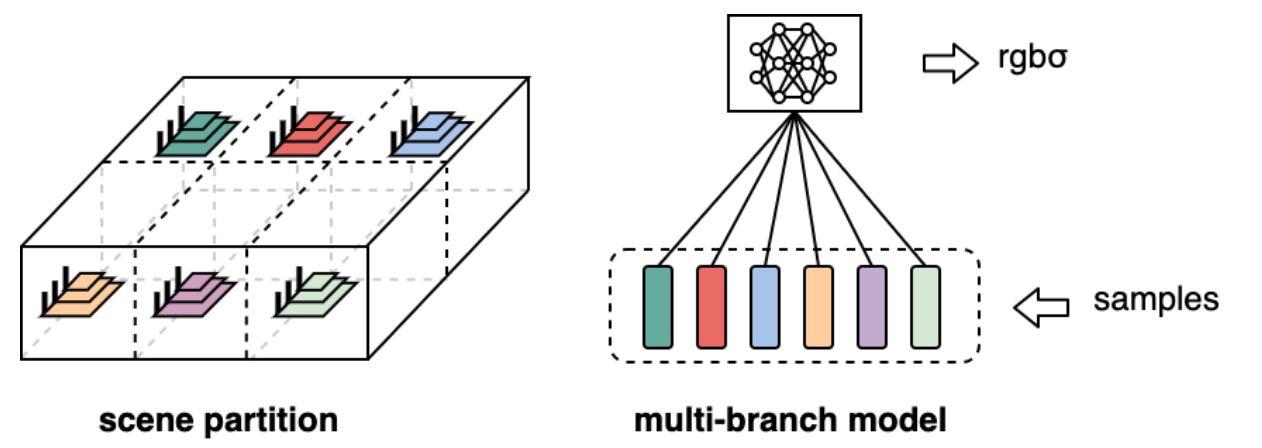}
    \caption[Multi-branch Model]{Multi-branch Model}
    \label{fig:enter-label}
\end{figure}
\begin{figure}
    \centering
    \includegraphics[width=0.5\linewidth]{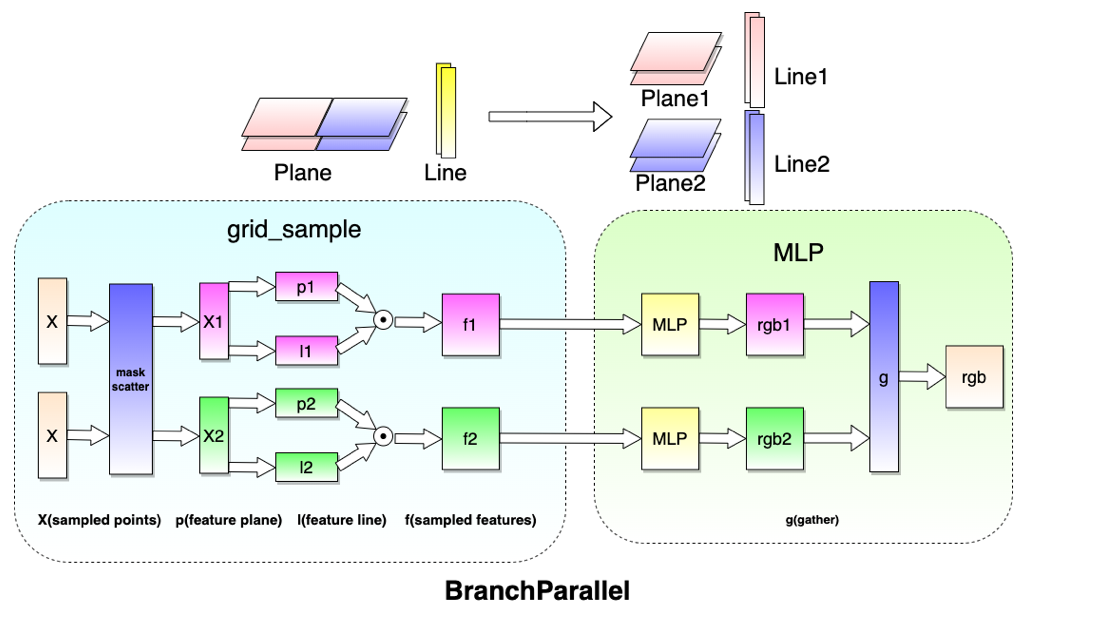}
    \caption[Branch Parallel]{Branch Parallel}
    \label{fig:enter-label}
\end{figure}


\begin{figure}[!h]
\centering
\includegraphics[width=0.8\columnwidth]{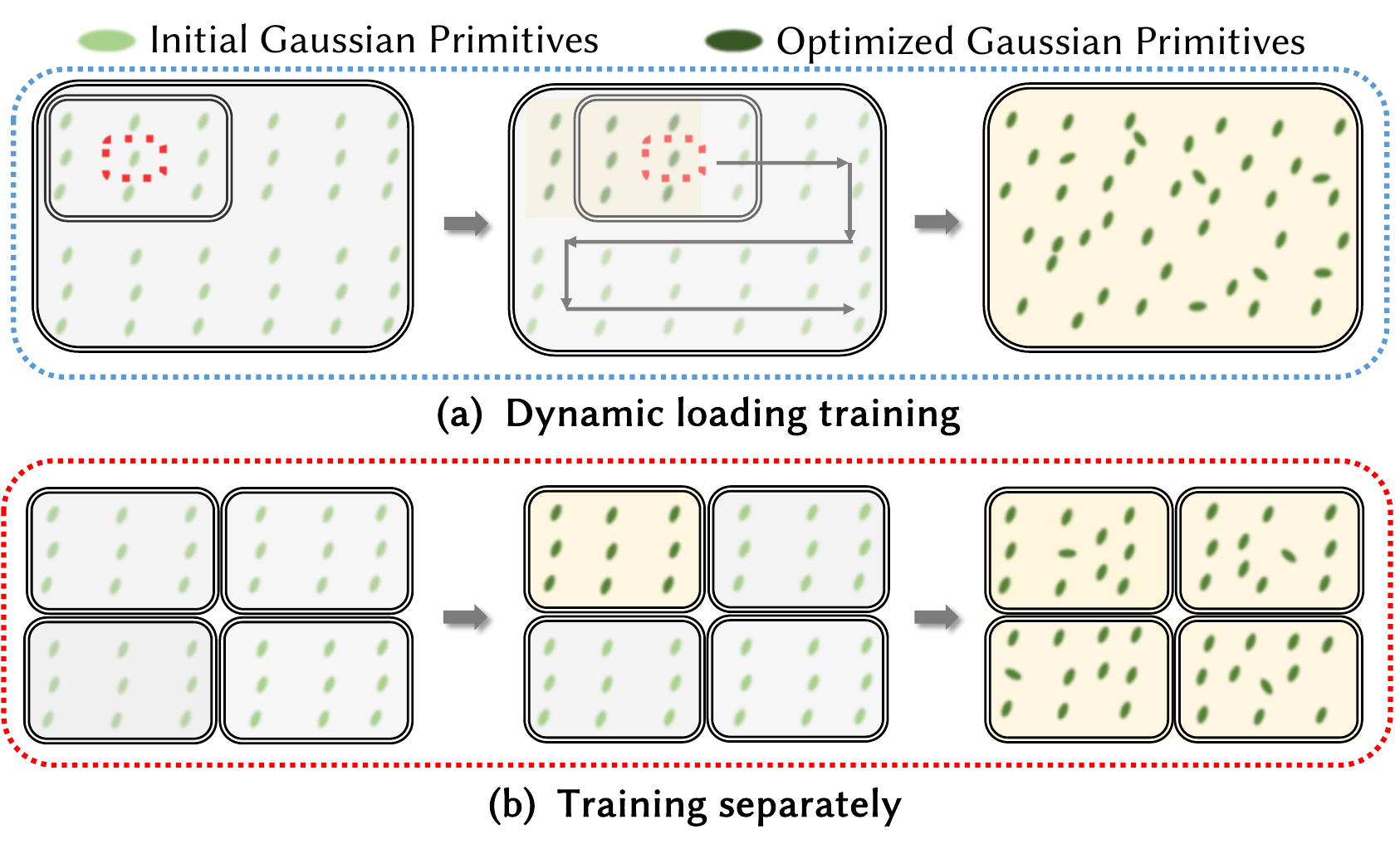} 
\caption[Dynamic loading Training]{Dynamic loading training vs. training each separate region independently. (a) Dynamic loading training optimizes local Gaussian primitives using a Zig-Zag trajectory for global synchronization. (b) The ‘Block-based’ method independently trains each region as a sub-mode
}
\label{fig_dyn_vs_block_training}
\end{figure}
\paragraph{Dynamic Loading Training}
For primitive based 3D Gaussain methods, GrendelGS\cite{zhao2024scaling3dgaussiansplatting} use pixel-wise and gaussian-wise distributed training to scale up. VastGS\cite{lin2024vastgaussian} partition the scene into several parts and train separately, which involves a lot of extra workloads for visual quality. Although model parallelism effectively reduces memory consumption, the additional distributed communication overhead significantly impairs scalability. Moreover, the requirement for multiple GPUs increases the barrier to entry for users in downstream tasks. In the context of oblique photography with a limited frustum, only a small subset of parameters is frequently utilized until the user moves to another area. Therefore, we propose dynamic loading to fully exploit the locality. Unlike model parallelism, dynamic loading can be performed on a single GPU, thereby reducing the hardware costs associated with training and rendering.

Our key insight is that the optimization of Gaussian primitives across the entire scene should be synchronized, which is crucial for eliminating seam artifacts. Fig. \ref{fig_dyn_vs_block_training} provides a visual comparison. Traditional block-based training methods often fail in maintaining a uniform update rate across different subregions. Even though parallel training provides some relief, Gaussian primitives at the boundaries typically receive inadequate supervision compared to their central counterparts. Therefore, rather than pre-partitioning the training images and optimizing subregions in isolation, we load and update a group of local Gaussian primitives at each iteration, and iteratively traverse the entire scene as the training progresses.

In practice, Gaussian primitives are initially stored in slower, lower-tier storage systems such as hard drives and CPU memory.  During each training iteration, a specific set of local Gaussian primitives, along with their corresponding optimizer states, is transferred to the faster memory of the GPU.  In the subsequent iteration, the previously optimized parameters are offloaded from the GPU memory and replaced by a new batch of parameters.  In this way, although the entire scene can be infinitely large, the size of a local region is bounded, consuming a restricted amount of GPU memory.  Eventually, all Gaussian primitives in the scene are fairly optimized with a balanced updating rate.

\begin{figure}[t]
\centering
\includegraphics[width=\columnwidth]{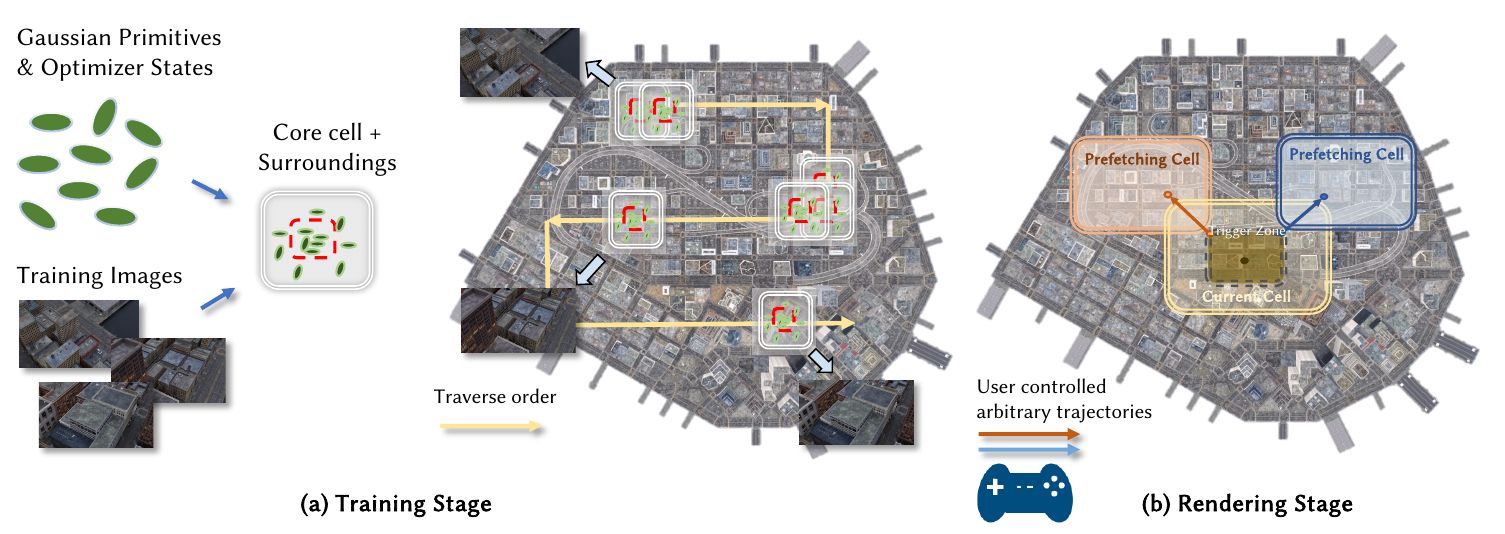} 
\caption[Dynamic loading Pipeline]{Dynamic loading pipeline for training and rendering large-scale 3D Gaussian primitives. Initially, all training data and Gaussian primitives are stored on CPU memory and disks. (a) During each training iteration, a set of local training images, along with their corresponding Gaussian primitives and optimizer states, are loaded onto the GPU for optimization. We follow a Zig-Zag trajectory to glance over the entire scene, with overlap between consecutive cells. (b) During rendering, we developed a prefetching strategy that uses a double-buffer to cache anticipating Gaussian primitives based on user’s moving direction and velocity. The trigger zone determines when and where to start prefetching. Once the user leave the trigger zone, the next Gaussian primitive cell will be loaded. Our system is able to stream smooth and seamless rendering results to user’s device even running on a single GPU.
}
\label{fig:pipline}
\end{figure}

\begin{figure}[t]
\centering
\includegraphics[width=0.6\columnwidth]{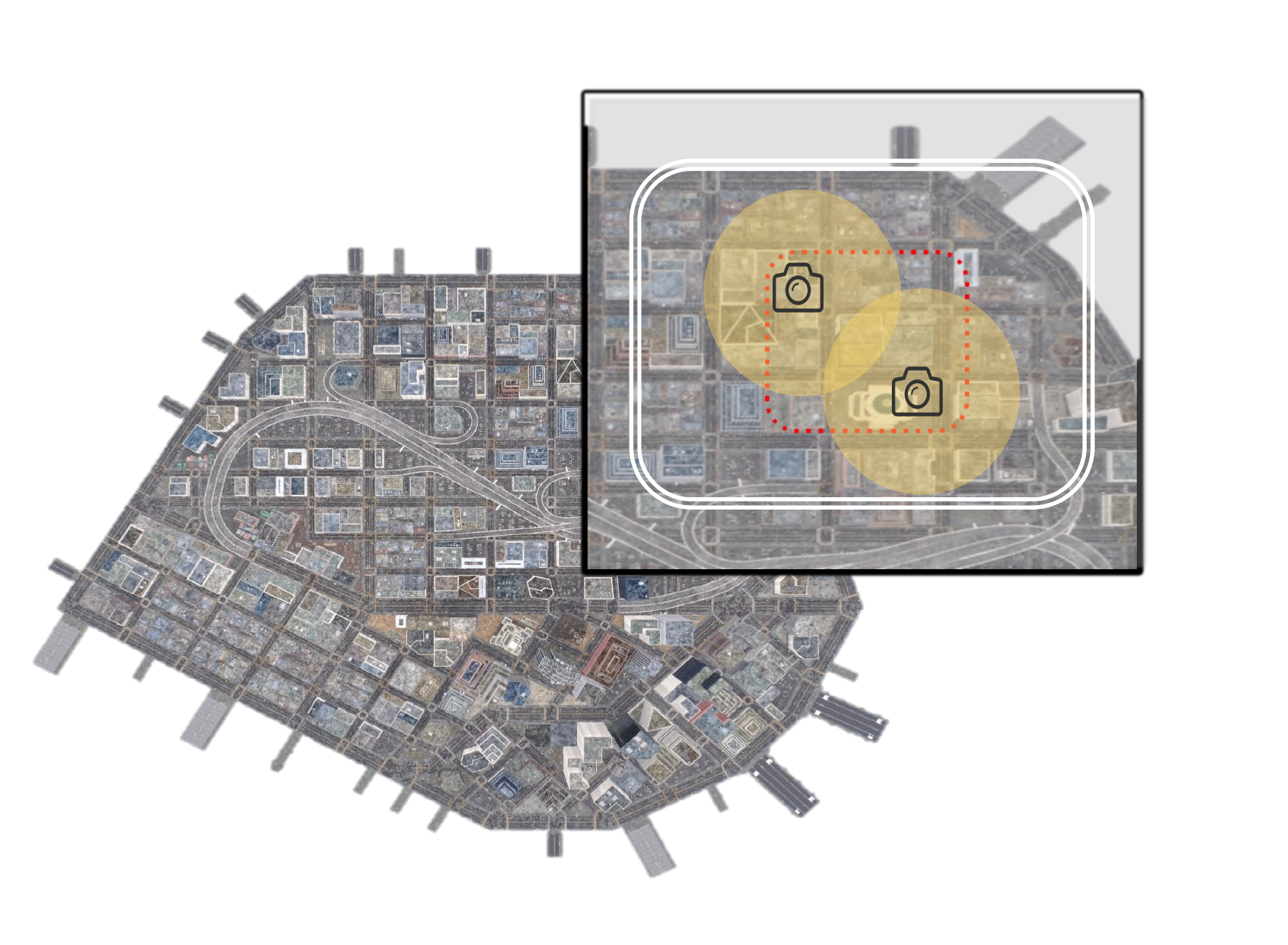} 
\caption[Scene Partition]{Diagram of the region of images and Gaussian primitives to onload during training. The red dashed box indicates the range of the core cell, and the remaining white dashed box is composed of the surrounding cells. Images whose camera centers fall within the core are applied to train the Gaussian primitives in both the core and surrounding cells.
}
\label{fig:cells}
\end{figure}

To reduce the communication overhead caused by frequent loading, it is preferable to utilize the data locality such that only partial parameters are loaded and offloaded instead of the entire batch. We therefore arrange training views and traverse the scene successively in a Zig-Zag pattern as illustrated in Fig. \ref{fig:pipline}.  Spatially close cameras are practically grouped together, and area covered by their frustums are treated as a cell.  We use such cell as the basic unit for dynamic loading.  The key is to ensure a certain amount of overlapping in both training views and Gaussian primitives between consecutively loaded cells.  We empirically found that, since cameras within a cell can look towards arbitrary directions, their frustums may extent beyond that cell.  Therefore, we enlarge the onloading range to also include the Gaussian primitives from surrounding cells, as illustrated in Fig. \ref{fig:cells}.

In summary, our dynamic loading training break the GPU memory barrier, towards large scale training on single device.The unique cell-based arrangement and Zig-Zag trajectory improve the training efficiency and avoid involving any extra computation.Previous works ~\cite{lin2024vastgaussian} concentrate on distributed training by block partition and add extra points to maintain the image quality, which may cause redundancy.Since ~\cite{lin2024vastgaussian} is not an open-source project, we implement it following its key insight and set as an additional choice for distributed training large scale scenes in LandMarkSystem.
\subsubsection{Accelerate Training in Parallel}
\label{DynLoad}
\label{DynLoad:Train}
Data parallel is widely used in DNN tasks for fast training speed. As the degree of parallelism of data parallelism increases, the model throughput can be expanded accordingly, thereby effectively improving training efficiency and accelerating convergence. Convolution neural networks different parts of dataset to different ranks and accumulate average gradients to update weights. For 3D reconstructions, the training process of GridNeRF is to input a batch of rays at a time for forward and reverse calculations. LandMarkSystem achieve data parallel by expanding and splitting the input rays.For methods based on 3DGS, training process requires densification for adaptive control, which depends on additional information about historical trained images to identify specific spatial zones to update. Additionally, only the 3D Gaussians included in the tile-based rasterization would be updated, which means that the parameters to update and densification strategy are not consistent across ranks, making naive data parallelism infeasible.

We experimented with naive data parallelism by accumulating and synchronizing average gradients across ranks, and then gathering and averaging results for densification. However, this approach resulted in a significant decline in image quality. Therefore, LandMarkSystem introduces a newly designed data pipeline mechanism to improve both training efficiency and image quality. Our analysis indicates that the quality degradation in naive data parallelism arises from a mismatch between densification requirements and gradient averaging. We address this by scheduling an early all-gather operation before backward to share loss values and other information. We then perform backward propagation with the aggregated loss and apply global-view densification. Experiments demonstrate that this optimized data parallel approach significantly accelerates training while preserving image quality.

\subsection{Techniques for Optimizing Rendering}
\subsubsection{Kernel Optimization} 

We conducted a function-level time analysis of the Landmark rendering stage and found that the \textbf{AnchorDecoder} and \textbf{RasterizeGaussians} operations accounted for approximately 43.2\% and 45.8\% of the rendering time, respectively, making them the main computational bottlenecks. 

\vspace{12pt}
\textbf{AnchorDecoder}
\vspace{6pt}

We conducted an operator-level fine-grained evaluation of the AnchorDecoder forward process. In typical rendering scenarios, the time distribution among the following components is as follows: \texttt{combine\_time} (tensor combination in parallel computation): 43\%, \texttt{mask\_time} (mask computation): 12\%, \texttt{cov\_time, color\_time, neural\_opacity\_time} (dual-layer MLP): 11\%, 8\%, 6\% respectively, \texttt{post\_time} (post-processing): 7\%, \texttt{cat\_local\_view\_wodist\_time} (local feature collation in multi-level parallelism): 6\%.
To ensure the effectiveness, portability, and adaptability of operator optimization, in addition to general optimizations like operator fusion and memory access consolidation, we have devised the following optimization methods for each step of the AnchorDecoder:
\begin{itemize}
\item  For the performance bottleneck combine\_time, we replaced the method of combining and then splitting tensors with fine-grained parallel computation, where each tensor is processed individually. This approach effectively reduces processing time, decreases memory usage, and improves speed.
\item As the second largest time-consuming component (approximately 25\%), integrating the dual-layer MLP can significantly accelerate the process. Additionally, the mask computation can be expedited by pre-processing the mask indices.
\item By pre-calculating the mask indices to reduce redundant calculations, the efficiency of the mask computation can be greatly improved.
\end{itemize}
The unit test of the optimized component AnchorDecoder shows a speedup of 2~3x. The end-to-end testing indicates that the forward process can output consistent and high-standard PSNR images, ensuring that user experience remains unaffected. The component supports various Gaussian Encoding algorithms, including VanillaGSEncoding, ScaffoldGSEncoding, and OctreeGSEncoding. The component has been deployed and validated on mainstream configurations such as NVIDIA A100, A800, and GeForce RTX 4090 GPUs, demonstrating high reliability and broad adaptability.

\vspace{12pt}
\textbf{RasterizeGaussians}
\vspace{6pt}

In addition to the optimization methods mentioned in the previous section, \texttt{RasterizeGaussians} also employs the strategies of  FlashGS~\cite{FlashGS} \footnote{Available at \href{https://github.com/InternLandMark/FlashGS}{https://github.com/InternLandMark/FlashGS}.} as follows. 

\begin{itemize}
\item Fuse the kernel functions preprocessCUDA and duplicateWithKeys. Since the former is a memory-intensive operator and the latter becomes a compute-intensive operator after inserting intersection operations for each tile, fusing these operators can achieve a more balanced compute-to-memory ratio and prevent a single factor from becoming an unoptimizable bottleneck.

\item Pass the transformation matrix shared by all threads via function argument passing, rather than using global memory, so that they are placed in constant memory, thereby reducing the bandwidth pressure on the L1 cache.

\item Independently schedule each warp to avoid synchronization overhead and also alleviate the problem of uneven load distribution among different warps within each block. 

\item Reduce stalls caused by memory access latency through the use of software pipelining techniques.

\item Reduce branch instructions, especially those that cause divergence among thread warps.

\item Reduce the number of executions of high-overhead operations and, within the allowable error margin, use inline assembly to reduce the overhead of exponential, logarithmic, and square root calculations.

\end{itemize}

\subsubsection{Dynamic Loading Rendering}

\label{DynLoad:Render}
\begin{figure}
    \centering
    \includegraphics[width=\columnwidth]{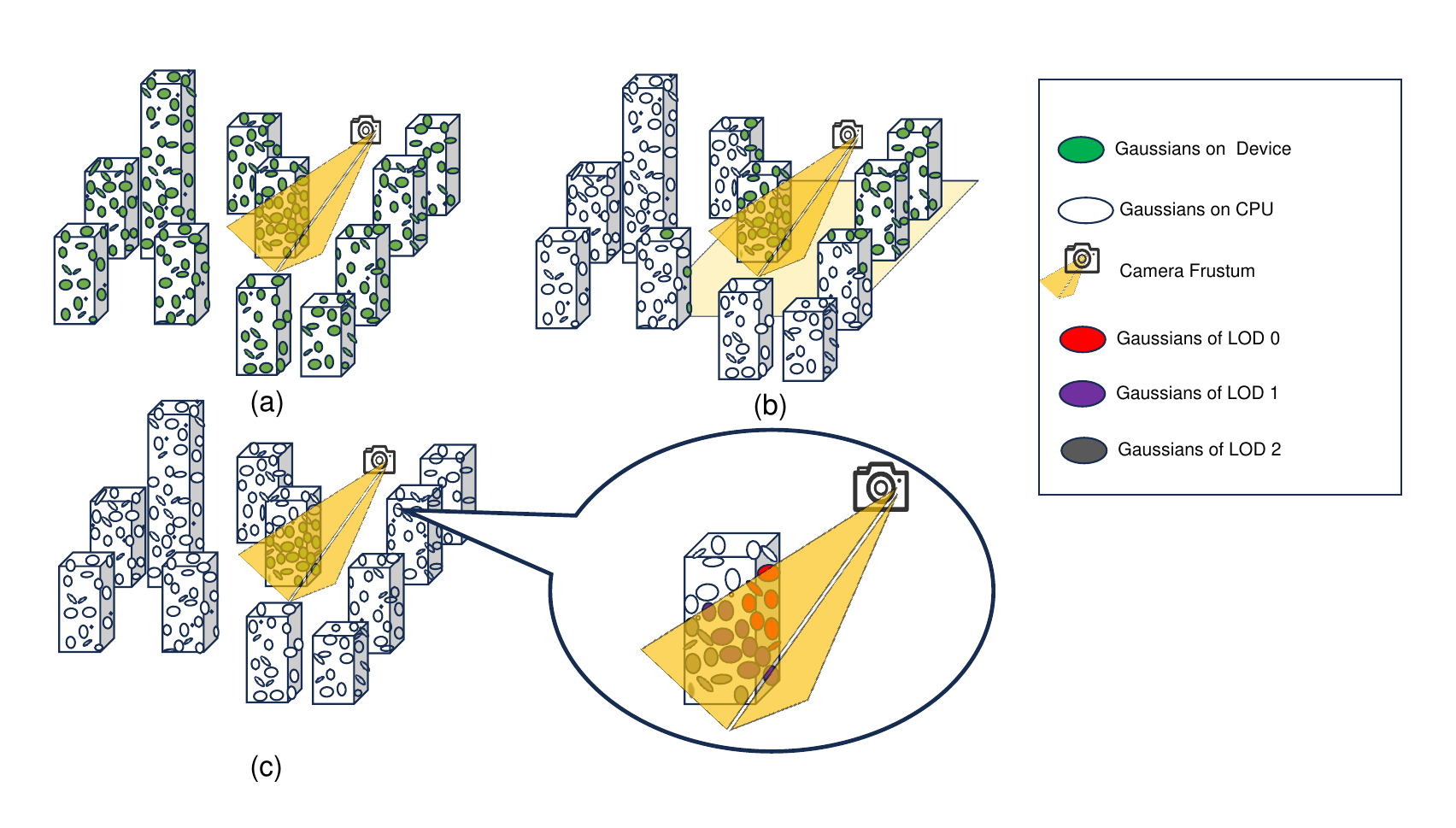}
    \caption{Three Methods for Image Rendering Using 3D Gaussian Splatting.(a) Without offload. (b) Block-based offload. (c) View-Frustum-based offload with LOD}
    \label{fig:3ways2render}
\end{figure}
For large scale scenes, the checkpoint is too large for modern compute devices. For instance, 3D Guassians for a \SI{25}{\kilo\meter\squared}
city scene in our experiment can take approximately 47GB memory, which largely exceeds the memory of consumer grade GPUs like NVIDIA RTX 4090.For NeRF methods, which usually require more parameters, it is much easier to reach the memory limit. Consequently, it is not practical to load all parameters once to a single GPU.  Distribute parameters to multiple GPUs like training phase could reduce the memory consumption per device, but is a waste of resources, ignoring the spatial locality and involving extra communications to synchronize. LandMarkSystem designs two novel rendering techniques adapted from parameter offloading, which is typically used for large language models\cite{rajbhandari2020zero}. The main difference is the granularity to group parameters. NeRF methods rely on parameters in MLP, while 3DGS uses local primitives that are flexible to partition. We have considered the compute graph based offloading technique, but found it hard to satisfy the strict demand on real-time rendering for 3D reconstruction. Inspired by computer graphics to cut the redundant elements out of sight, we designed both block-based and frustum-based methods for seamless viewing large-scale on memory constrained devices. 

\paragraph{Block-based dynamic loading rending}
Partition and load a part is natural in computer graphics. The challenge with rendering lies in the necessity to deliver real-time rendering. A simple loading process can result in noticeable wait times, which negatively impacts the user experience. In the context of dynamic loading, prefetching parameters are often used to overlap the loading time. However, in scenarios that allow for free preview, unlike sequential computations, the user’s subsequent position is uncertain, making it difficult to prefetching in advance. To achieve seamless browsing, we have designed a block based loading method based on user intention prediction, utilizing double-buffering to ensure that loading does not influence the current rendering, thereby allowing the two processes to overlap.In the context of the NeRF method, conventional slicing does not effectively leverage spatial locality. We employ the previously mentioned multi-branch structure to spatially segment the model, thereby enabling the block-based dynamic loading approach to be applicable to NeRF as well.

To be specific, we implemented a prefetching strategy that utilizes double-buffer to temporarily store parameters, effectively balancing memory usage against speed. Specifically, one buffer holds the currently observed cell, while the other buffer prefetches the cell that is about to come into view. Determining when and where to prefetch is crucial. As depicted in Fig. \ref{fig:pipline} (b), we setup a trigger zone for prefetching, which is a bounded region enclosed by the currently loaded cell. Within the rendering scene, the user periodically sends a new displacement value, altering their position and viewing angle. We have designed a dual loading area strategy. When the user’s displacement extends beyond the first area, the backend loading buffer initiates the loading of parameter regions mapped to the direction of movement. As the user moves beyond the second area, the rendering buffer is switched to the preloaded buffer. The size of each area is meticulously designed to match the user’s movement speed and the hardware’s loading efficiency. This process is quick and seamless, virtually imperceptible to the human eye. This ensures that the user’s view frustum, pointing in any direction, will have both cells readily available, thereby maintaining a smooth visual experience.

It is worth noting that our dynamic loading strategy utilizes a consistent amount of memory, which correlates with the size of the cell being loaded. This allows us to tailor our approach to various computing devices by modifying the cell size. In essence, we load smaller regions of Gaussian primitives for the device with limited memory capacity, and conversely, larger segments for devices with more memory available. 

\paragraph{Frustum-based dynamic loading rending with LOD}
While block-based dynamic loading method is suitable for both NeRF and 3DGS, it constraint limits the user’s ability to view distant areas, as doing so would require the use of larger block sizes, which could potentially lead to insufficient GPU memory. We further propose a view-frustum-based dynamic loading rendering approach that leverages the independence of Gaussian spheres in the 3D Gaussian splitting algorithm. The view-frustum-based method aligns more closely with human visual perception, as it only loads the content within the field of view, while omitting the parts outside the view from being loaded onto the GPU. This approach can further reduce memory usage compared to both static loading and block-based dynamic loading. In static loading, all Gaussian spheres are loaded onto the GPU, while in block-based dynamic loading, all Gaussian spheres within the currently rendered block of space are loaded. In both cases, the number of Gaussian spheres loaded exceeds that of the view-frustum-based dynamic loading method. The range of Gaussian spheres loaded by the three rendering methods is shown in Fig.\ref{fig:3ways2render}.

Furthermore, another significant advantage of employing a view-frustum-based dynamic loading rendering approach is its ability to facilitate six degrees of freedom (6-DOF) viewpoint changes within the spatial domain without the concern of encountering blank regions devoid of loaded Gaussian spheres. This method operates on the principle of loading Gaussian spheres within the view frustum into the GPU for rendering. Consequently, even when observing from level or other equiangular perspectives, it ensures that the view frustum contains the Gaussian spheres intended for display. Such viewing angles cannot be effectively supported by block-based dynamic loading rendering methods, which are limited to providing only a restricted range of overhead viewing perspectives. This can also be observed in Fig.\ref{fig:3ways2render}(b), where the range of the block area to be loaded is determined based on the current camera space position. However, when the camera view is not a downward overhead perspective but instead looks towards an area outside the block region, areas without loaded Gaussian spheres become visible, resulting in blank spaces.

To achieve the desired effect, the trained Gaussian spheres must be reordered based on their spatial arrangement, a process performed offline. The purpose of reordering is to assign the previously unordered Gaussian spheres to their respective three-dimensional voxel grids, which are organized sequentially to enable fast indexing of the contained Gaussian spheres. Afterward, a view-frustum filtering process is applied to the voxel grids to identify those within the visible range of the view frustum, significantly reducing the computational burden associated with view-frustum filtering.

From a performance perspective, while frustum-based loading may occur frequently, the overall computational complexity is optimized through efficient indexing and reduced mask sizes. This optimization compensates for the overhead introduced by loading, ensuring that the dynamic loading of the visible cone meets performance requirements. Additionally, to minimize the time spent loading Gaussian spheres into the fixed GPU cache, asynchronous loading is employed. This approach allows the Gaussian spheres needed for the next frame’s view frustum to be loaded in parallel while the current frame is being rendered, effectively hiding the additional time overhead associated with transferring Gaussian spheres to the GPU.

\section{Evaluation} \label{sec:criteria}
Our experiments aim to validate the capabilities of our system in both training and rendering scenarios.

\subsection{Training}
\label{sec:eva_training}
We utilized a subset of the MatrixCity dataset~\cite{li2023matrixcity}, comprising 1063 images for training and 163 images for testing, all with a resolution of $1920 \times 1080$. We evaluated the effectiveness of our training optimizations for NeRF and 3DGS models as shown in Tab.~\ref{tab:nerf-gs-comparison}.

\begin{table}[htbp]
    \centering
    \setlength{\tabcolsep}{8pt} 
    \renewcommand{\arraystretch}{1.2} 
    
    \begin{tabular}{ccccccc}
        \toprule
        Devices & 1 & 2 & 4 & 8 & 16 \\ 
        \midrule
        Speedup & 1.00 & $1.60\times$ & $2.76\times$ & $4.86\times$ & $9.82\times$ \\ 
        PSNR & 25.06 & 24.95 & 24.64 & 24.53 & 24.52 \\ 
        Primitives & 3.7M & 3.7M & 3.6M & 3.4M & 2.7M \\ 
        Memory(GB) & 5.99 & 6.89 & 6.87 & 6.41 & 5.25 \\ 
        \bottomrule
    \end{tabular}
    \caption{Data parallel performance of 3DGS on different number of devices.}
    \label{tab:data_parallel_results}
\end{table}

\begin{table}[h]
\centering
\begin{tabular}{>{\raggedright\arraybackslash}p{5.5cm} @{\hspace{15pt}} r @{\hspace{20pt}} >{\raggedright\arraybackslash}p{5.5cm} @{\hspace{5pt}} r}
\toprule
\textbf{NeRF Model} & \textbf{PSNR} & \textbf{3DGS Model} & \textbf{PSNR} \\
\midrule
GridNeRF                  & 25.021 & Vanilla GS                     & 24.988 \\
GridNeRF (Data Parallel)    & 24.707 & Vanilla GS (Data Parallel)       & 24.494 \\
GridNeRF (Channel Parallel)    & 25.137 & Vanilla GS (Dyn)      & 24.268 \\
GridNeRF (Branch Parallel)    & 25.022 & Scaffold GS               & 27.740 \\
Instant-NGP             & 24.331 & Scaffold GS (Data Parallel)  & 27.175 \\
Instant-NGP (Data Parallel) & 24.184 & Scaffold GS (Dyn)  & 27.111 \\
Instant-NGP (Branch Parallel) & 24.598 & Octree GS                 & 27.765 \\
Nerfacto                     & 24.866 & Octree GS (Data Parallel)       & 27.512 \\
Nerfacto (Data Parallel)        & 24.947 & Octree GS (Dyn)       & 27.250 \\
Nerfacto (Branch Parallel)        & 24.971 & & \\
\bottomrule
\end{tabular}
\caption{Comparison of NeRF and 3DGS Models}
\label{tab:nerf-gs-comparison}
\end{table}

Channel Parallel and Branch Parallel are the model parallel implementations we proposed for the NeRF Models to address the issue of insufficient GPU memory for large-scale training. When combined with data parallelism, the training convergence time can be significantly reduced given sufficient resources. From the final convergence situation, the final accuracy difference between single-card and parallel training is within $3\%$.

For large-scale scene training with 3DGS, we implemented a combination of dynamic training and data parallelism, which differs from model parallelism by offering higher resource utilization efficiency. Dynamic loading training enables the training of unlimited areas on a single GPU, while data parallelism accelerates training speed when hardware resources are sufficient.

The results in Tab.~\ref{tab:nerf-gs-comparison} confirm the effectiveness of data parallelism and dynamic loading for GS algorithms. The final convergence errors of models trained with these methods are within $3\%$ of those trained on a single card. Since Scaffold GS and Octree GS continue training on point clouds generated by Vanilla GS, their final accuracy is higher than other algorithms.

\subsection{Rendering}

We also evaluated the rendering performance of our optimizations for NeRF and 3DGS models.

\subsubsection{NeRF}
We first validated the effect of hybrid parallelism on GridNeRF, as shown in Tab.~\ref{tab:gridnerf-parallelism-comparison}.

\begin{table}[h]
    \centering
    \begin{tabular}{lcc}
        \toprule
        \textbf{NeRF Model} & \textbf{Latency (s)} & \textbf{GPU Mem (GB)} \\
        \midrule
        GridNeRF & 2.022 & 4.223 \\
        GridNeRF (4 DP) & 0.540 & 3.491 \\
        GridNeRF (8 DP) & 0.279 & 3.436 \\
        GridNeRF (2 TP x 2 DP) & 2.932 & 2.812 \\
        GridNeRF (2 TP x 4 DP) & 1.515 & 2.812 \\
        \bottomrule
    \end{tabular}
    \caption{Performance of different parallelism strategies in GridNeRF. DP stands for data parallelism and TP stands for tensor parallelism.}
    \label{tab:gridnerf-parallelism-comparison}
\end{table}

We measured the average end-to-end latency for rendering an image and the peak memory usage per GPU during the rendering process. While 2 degree tensor parallelism (TP) reduces memory usage by $1.5\times$, it increases average latency by $1.45\times$ due to communication overhead. Data parallelism (DP) effectively reduces latency, with performance improvement nearly linearly scaling with the number of GPUs.

\begin{table}[h]
    \centering
    \begin{tabular}{lcc}
        \toprule
        \textbf{NeRF Model} & \textbf{Latency (s)} & \textbf{GPU Mem (GB)} \\
        \midrule
        GridNeRF & 4.696 & 62.709 \\
        GridNeRF (4 DP) & 1.266 & 62.519 \\
        GridNeRF (8 DP) & 0.649 & 62.436 \\
        GridNeRF (offload) & 2.790 & 25.119 \\
        GridNeRF (offload + 4 DP) & 0.778 & 24.935 \\
        GridNeRF (offload + 8 DP) & 0.421 & 24.854 \\
        \bottomrule
    \end{tabular}
    \caption{Performance of data parallelism and offloading (dynamic loading) in GridNeRF.}
    \label{tab:gridnerf-offload-comparison}
\end{table}

We defined a movement trajectory in a larger model which is 25$\times$ the size of the above model and collected the average latency and peak memory usage of the movement. As shown in Tab.~\ref{tab:gridnerf-offload-comparison}, in the case of a single card, the average rendering latency is 4.696s, and the average performance improvement is 3.71$\times$ and 7.24$\times$ in the case of 4 DP and 8 DP, respectively. After adding offload, the performance of single-card inference is not only improved by 1.68$\times$, but also the memory usage is reduced from the original 62.709 GB to 25.119 GB, a reduction of 2.5$\times$.

\subsubsection{3DGS}

For 3DGS algorithms, we directly compared the performance of the original open-source implementation and the LandMarkSystem implementation based on the models trained in Section~\ref{sec:eva_training}. The results are presented in Tab.~\ref{tab:3DGS-performance-comparison}.

\begin{table}[h]
    \centering
    \begin{tabular}{lcc}
        \toprule
        \textbf{GS Model} & \textbf{Latency (ms)} & \textbf{FPS} \\
        \midrule
        Vanilla GS (open source) & 15.53 & 64.374 \\
        Vanilla GS (LandMarkSys) & 4.56 & 219.263 \textcolor{blue}{(3.41$\times$)} \\
        Scaffold GS (open source) & 9.39 & 106.532 \\
        Scaffold GS (LandMarkSys) & 5.32 & 187.957 \textcolor{blue}{(1.76$\times$)} \\
        Octree GS (open source) & 17.401 & 57.467 \\
        Octree GS (LandMarkSys) & 10.01 & 99.894 \textcolor{blue}{(1.73$\times$)} \\
        \bottomrule
    \end{tabular}
    \caption{Performance comparison of 3DGS models: original open-source implementation vs. LandMarkSystem implementation on a single GPU.}
    \label{tab:3DGS-performance-comparison}
\end{table}

For Vanilla GS, we have inference parameter cache, frustum culling and rasterizer kernel optimization. So its improvement in terms of FPS is the highest, up to 3.41$\times$. For Scaffold GS and Octree GS, our rasterizer kernel optimization can also have an effect on them. In addition, we also have fused anchor decoder optimization, so their performance improvement is 1.76$\times$ and 1.73$\times$ respectively.

We extended the experiments to a larger scene, MatrixCity-Small, to demonstrate the generality of our system optimizations. The results, shown in Tab.~\ref{tab:3DGS-performance-comparison-large}, indicate that our optimizations are effective even in larger scenes. Vanilla GS, Scaffold GS, and Octree GS in the LandMarkSystem exhibit FPS improvements of 5.92$\times$, 1.57$\times$, and 1.33$\times$, respectively, compared to the original open-source implementation. Furthermore, the LandMarkSystem's implementation is more GPU memory-efficient during rendering.

\begin{table}[h]
    \centering
    \begin{tabular}{lccc}
        \toprule
        \textbf{GS Model} & \textbf{Latency (ms)} & \textbf{FPS} & \textbf{GPU Mem (GB)} \\
        \midrule
        Vanilla GS (open source) & 32.748 & 30.536 & 11.602 \\
        Vanilla GS (LandMarkSys)& 5.528 & 180.895 \textcolor{blue}{(5.92$\times$)} & 6.234 \\
        Scaffold GS (open source) & 14.849 & 67.341 & 7.874 \\
        Scaffold GS (LandMarkSys) & 9.429 & 106.048 \textcolor{blue}{(1.57$\times$)} & 7.642 \\
        Octree GS (open source) & 17.295 & 57.817 & 5.463 \\
        Octree GS (LandMarkSys) & 12.994 & 76.957 \textcolor{blue}{(1.33$\times$)} & 5.438 \\
        \bottomrule
    \end{tabular}
    \caption{Performance comparison of 3DGS models in MatrixCity-Small: original open-source implementation vs. LandMarkSystem implementation on a single GPU.}
    \label{tab:3DGS-performance-comparison-large}
\end{table}

\begin{figure}
    \centering
    \includegraphics[width=1.0\linewidth]{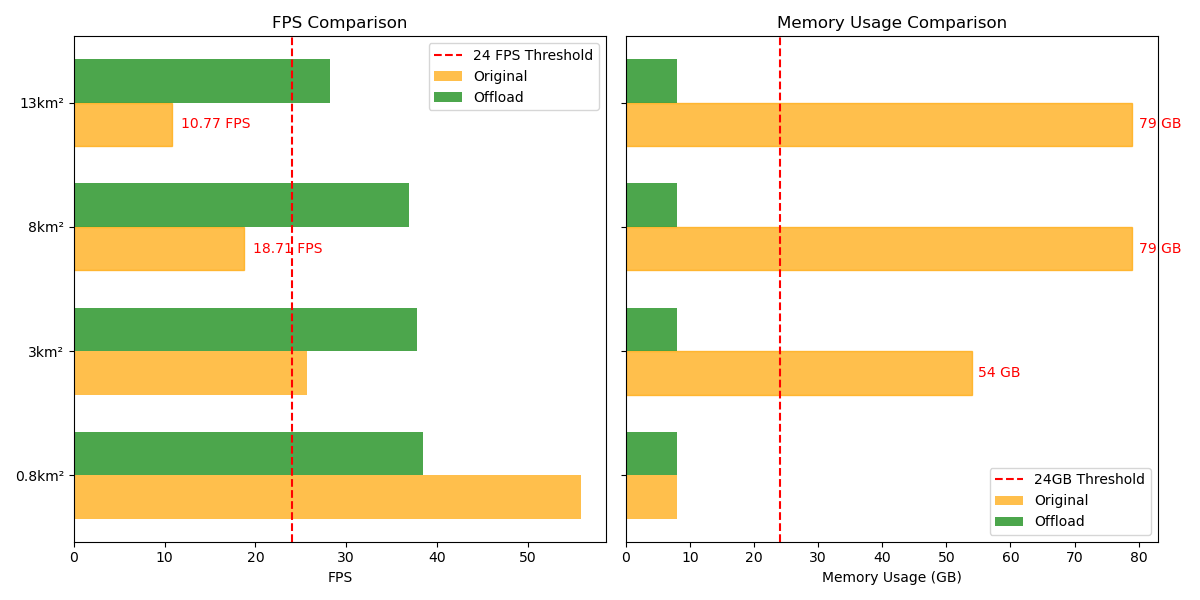}
    \caption[Offload-data]{Comparison between frustum-based offload and original method with OctreeGS.The 24 FPS threshold represents the real-time threshold. The 24GB threshold represents the memory of NVIDIA RTX4090, which is currently the best of the consumer-grade video card.}
    \label{fig:offload_data}
\end{figure}

To showcase the impact of the offloading method, we employ the Octree Gaussian algorithm as the foundational approach and evaluate its performance in rendering scenes of varying sizes on a single NVIDIA A100. As depicted in Fig. \ref{fig:offload_data}, the checkpoint size grows with the scene size, leading to the failure of the original Octree Gaussian algorithm to render scenes on consumer-grade GPUs due to memory limitations. In contrast, our view-frustum-based 3D Gaussian dynamic loading rendering method keeps GPU memory usage stable and constant while ensuring real-time rendering performance.

It is important to note that the offloading method may perform slower than the original approach due to the additional loading processes. However, in cases where memory consumption is low, offloading is unnecessary. For larger scenes, our spatial fast indexing significantly accelerates sorting, achieving fast rendering speeds while maintaining stable and minimal memory usage.

\section{Related Works}
\subsection{Neural Rendering Frameworks}
Shortly after the emergence of NeRF, a plethora of NeRF algorithm variants surfaced, including Mip-NeRF\cite{9710056}, Plenoxels\cite{9880358}, and TensoRF\cite{10.1007/978-3-031-19824-3_20}.  These variants have unique algorithm structures, lacking a universal computation process, which obstacles the utilization of different optimization strategies. Consequently, several systematic frameworks have been proposed to standardize the computational pipelines for various NeRF structures. NeRF-Factory\cite{NeRF-Factory} collects existing NeRF works to share the same input and output formats for benchmarking. It is not designed to build reusable components between the algorithm structures. NerfAcc\cite{10376856} modularizes NeRF computation process into Python modules and accelerates relevant pipeline stages (e.g., samplers for space sampling). It lacks a comprehensive and unified pipeline atop these independent components. Kaolin-wisp\cite{KaolinWispLibrary} achieves feasibility both in modularizing computation processes and unifying pipelines. Its reusability is limited to specific NeRF structures. Building upon previous work, Nerfstudio\cite{nerfstudio} implements a modular and unified NeRF computational framework that supports real-time visual rendering for user-captured real-world scenes. It primarily focus on algorithmic modification and iteration instead of  the bottlenecks in system resources, making it not suitable for large-scale scene training and rendering tasks. LandMarkSystem, from the perspective of NeRF computational systems, provides fully modular model and pipeline components for different algorithm structures. It adapts to various real-world scene scales in reconstruction and supports real-time rendering of  the scene up to hundreds of square kilometers

\subsection{Large-scale Scene Reconstruction}
As a novel representation method for 3D models, NeRF has garnered widespread attention for its ability to capture fine details and reproduce realism beyond previous methods, especially in large-scale scenes. With increasing scene complexity and information density, optimization techniques have emerged. BungeeNeRF\cite{10.1007/978-3-031-19824-3_7} proposes a progressive model growing and training method to adapt to multi-scale datasets, from satellite height to individual buildings. The number of residual modules in the model increases with the scale levels. MegaNeRF\cite{9878491} segments large regions, with each subregion handled by a single NeRF model. It predicts the sample color and density along the ray based on the corresponding model, followed by volumetric rendering to obtain the final pixel color. BlockNeRF\cite{9879943} Validates this method further on large-scale street view images, constructing real-world scene models that allow local replacement and updates. However, this training paradigm becomes intensive both in storage and computation as the region size increases, making it challenging to scale to even larger areas. LandMarkSystem Addresses this scalability issue by proposing multiple parallel training strategies applicable to various NeRF algorithm structures. After the emergence of the 3DGS method, research on applying it to large-scale scene reconstruction rapidly proliferated. CityGaussian\cite{liu2024citygaussianrealtimehighqualitylargescale} Introduces a training and level-of-detail (LOD) generation method based on scene partitioning. This enables parallel training along region dimensions under 3DGS, enhancing rendering speed by blending different LOD scene blocks based on distance. RetinaGS\cite{li2024retinagsscalabletrainingdense} Also employs scene partitioning for parallel training and reconstruction in large-scale scenes. It allocates relevant computational tasks according to 3D Gaussian positions, achieving load balancing between multiple GPUs during the training and rendering processes. GrendalGS\cite{zhao2024scaling3dgaussiansplatting}  Utilizes the sparse properties of 3D Gaussian distributions, employing a different model splitting strategy. It randomly distributes 3D Gaussians across parallel GPUs, achieving storage and computation balance during training and significantly improving large-scale scene reconstruction quality. LandMarkSystem Specifically addresses large-scale scene reconstruction in 3DGS with Dynamic Loading Training. It supports training ultra-large  models on a single GPU and can be combined with distributed data parallelism to enhance data throughput during training.

\subsection{Large Model Parameters Offload}
The increasing scale of the scene leads to explosive growth in the parameters of the reconstruction model. This correlation is similar to the scaling laws\cite{kaplan2020scalinglawsneurallanguage} of language model parameters. To address the challenges posed by the increase in model parameters on limited system resources, related work focuses on achieving horizontal and vertical scaling of model training and inference. Horizontal scaling training methods include Megatron-LM\cite{10.1145/3458817.3476209}, which splits the model parameters into several independent parts, and Gpipe\cite{10.5555/3433701.3433727}, which divides the model's layer-by-layer computations into multiple pipeline stages, and then distributes them across multiple GPU devices for parallel computation. ZeRO\cite{10.5555/3433701.3433727} proposes a partitioning method different from the above perspectives, avoiding the  redundancy in parameters, gradients, and optimizer states across all parallel devices through collective communication, further reducing the demand on system storage resources. Vertical scaling training methods aim to increase performance on a single device. ZeRO-Offload\cite{273920} proposes a method that cooperatively utilizes CPU storage, by offloading and loading  the model parameters, to train models that far exceed the GPU memory capacity. During model inference, horizontal scaling methods based on model partitioning can also be used. ZeRO-Inference\cite{aminabadi2022deepspeedinferenceenablingefficient} proposes a method that integrates NVMe and CPU storage to offload and load model parameters, reducing resource usage during the inference stage through vertical scaling, and improving the feasibility of deploying models on limited system resources. LandMarkSystem not only provides horizontal scaling optimizations for NeRF and 3DGS both in training and rendering but also offers vertical scaling strategies based on dynamic loading. Unlike existing methods, LandMarkSystem designs parameter dynamic management components based on the spatial distribution characteristics of model parameters in large-scale scene reconstruction tasks, further implementing spatially-based dynamic loading training and rendering optimization strategies. 

\section{Conclusion}
The LandmarkSystem framework constructs a comprehensive 3D reconstruction system that is not only suitable for casual-scale scenarios but also provides multiple optimization techniques for super large-scale scenarios, making super large-scale 3D reconstruction more feasible. LandmarkSystem includes current algorithm components related to NeRF (Neural Radiance Fields) and 3DGS (3D Gaussian Splatting), along with a Converter that facilitates the transformation of models into distributed models. With these components, people who want to create a new NeRF or 3DGS model can easily build one. Besides, people can converter that model to distributed with few effort and get efficient training and rendering experience. In addition, the cleverly designed Dynamic Loading module can greatly reduce the amount of GPU memory required for training and rendering. Using Dynamic Loading can achieve unlimited area training and rendering with limited computing resources. LandmarkSystem implements a unified training and rendering pipeline, ensuring that models trained within LandmarkSystem can be directly used for distributed rendering deployment. These features collectively enhance the usability and efficiency of the system in both casual and large-scale scenarios. Some evaluation experiments are carefully designed and show that LandmarkSystem can train super large-scale model without PSNR loss and extremely up-speeds rendering even with much less GPU memory.

\bibliographystyle{unsrt}
\bibliography{main}

\end{document}